\documentclass[10pt,twocolumn,letterpaper]{article}

\usepackage{cvpr}
\usepackage{times}
\usepackage{epsfig}
\usepackage{graphicx}
\usepackage{amsmath}
\usepackage{amssymb}

\usepackage{algorithm}
\usepackage{algpseudocode}
\usepackage{multirow}
\usepackage{blindtext}
\usepackage{array, booktabs}
\usepackage{arydshln}
\usepackage{calrsfs}
\usepackage{subcaption}
\usepackage{appendix}
\usepackage{lipsum}
\usepackage{multicol}

\usepackage{kotex}
\usepackage{color}
\DeclareMathAlphabet{\pazocal}{OMS}{zplm}{m}{n}

\definecolor{greyblue}{rgb}{0.1,0.6,0.5}

\definecolor{purple}{rgb}{0.626,0.125,0.941}

\usepackage[pagebackref=true,breaklinks=true,letterpaper=true,colorlinks,bookmarks=false]{hyperref}

\cvprfinalcopy 

\def\cvprPaperID{4706} 

\ifcvprfinal\fi
\begin{document}

\title{TedEval: A Fair Evaluation Metric for Scene Text Detectors}

\author{
 Chae Young Lee\footnotemark[1] , Youngmin Baek\thanks{Authors contributed equally.} , and Hwalsuk Lee\thanks{Corresponding author.}\\
 Clova AI Research, NAVER Corp.\\
 {\tt\small\{cylee.ai, youngmin.baek, hwalsuk.lee\}@navercorp.com} \\
}

\maketitle

\begin{abstract}
Despite the recent success of scene text detection methods, common evaluation metrics fail to provide a fair and reliable comparison among detectors. They have obvious drawbacks in reflecting the inherent characteristic of text detection tasks, unable to address issues such as granularity, multiline, and character incompleteness. In this paper, we propose a novel evaluation protocol called TedEval (Text detector Evaluation), which evaluates text detections by an instance-level matching and a character-level scoring. Based on a firm standard rewarding behaviors that result in successful recognition, TedEval can act as a reliable standard for comparing and quantizing the detection quality throughout all difficulty levels. In this regard, we believe that TedEval can play a key role in developing state-of-the-art scene text detectors. The code is publicly available at \href{https://github.com/clovaai/TedEval}{https://github.com/clovaai/TedEval}.
\end{abstract}

\section{Introduction}
Along with the progress of deep learning, the performance of scene text detectors have remarkably advanced over the past few years~\cite{tian2016ctpn, zhou2017east, shi2017seglink}. However, providing a fair comparison among such methods is still an ongoing problem, for common metrics fail to reflect the intrinsic nature of text instances~\cite{dangla2018eval, liu2019tiou}. One example is the IoU (Intersection over Union) metric~\cite{everingham2015pascal}. Adopted from the object detection Pascal VOC, it is designed to detect bounding boxes containing a single object and thus is not suitable to detect text instances consisting of multiple characters. Another approach is DetEval, a metric specifically designed to evaluate text bounding boxes~\cite{wolf2013deteval}. However, its unreasonably lenient criteria accepts incomplete detections that are prone to fail in the recognition stage. Recently, Liu et al has proposed a new metric named TIoU (Tightness-aware IoU), which adds the tightness penalty to the IoU metric~\cite{liu2019tiou}. This approach can be problematic in that it relies on the quality of the Ground Truth (GT), which is often inconsistent.


\begin{figure}[t]
    \begin{subfigure}{.5\linewidth}
    \centering
    \includegraphics*[width=\linewidth, height=6.5cm, clip=true]{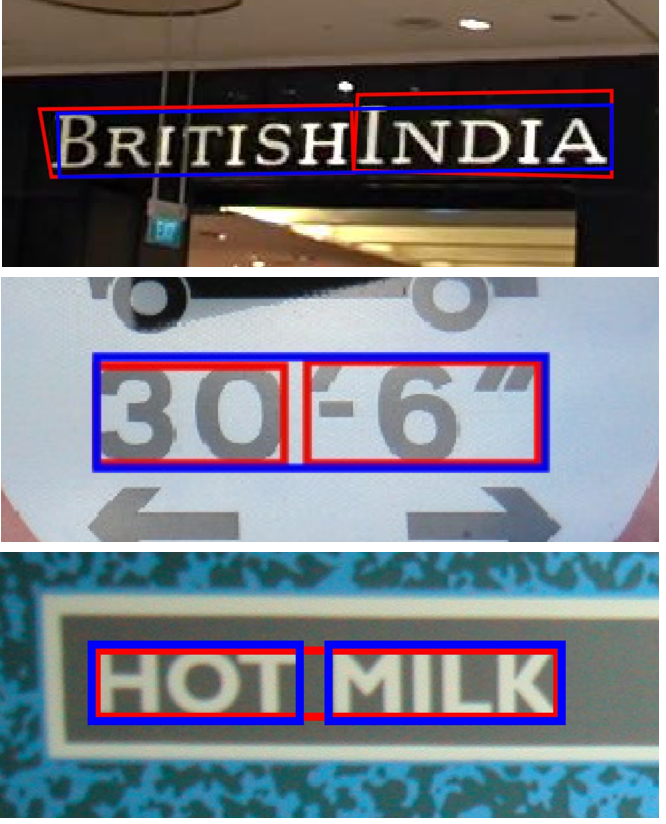}
    \caption{Granularity (IoU=0.0)}
    \end{subfigure}%
    \begin{subfigure}{.5\linewidth}
    \centering
    \includegraphics*[width=\linewidth, height=6.5cm, clip=true]{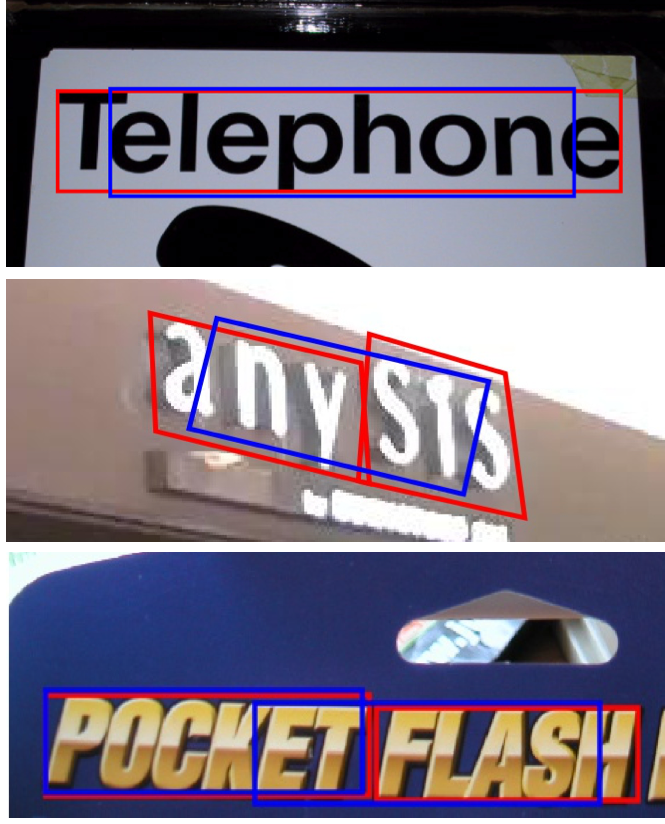}
    \caption{Completeness (DetEval=1.0)}
    \end{subfigure}
    \vspace{-3mm}
    \caption{Examples of unfair evaluations. (a) rejected by IoU but should be accepted. (b) accepted by DetEval but should be penalized. Red: GT. Blue: detection.}
  \label{fig:problem}
  \vspace{-3mm}
\end{figure}

To solve these issues, a fair evaluation metric for text detectors must account for:

\begin{itemize}
    \item {\bf Granularity}
    Annotation tendencies of public datasets vary greatly due to the lack of gold standard on bounding box units. Merging and splitting detection results to match GT should be allowed to a reasonable extent.
    \item {\bf Completeness}
    An instance-level match may accept incomplete detections that have missing or overlapping characters. A penalty must be given so that scores reflect whether the detection contains all required characters.
\end{itemize}

In the proposed metric called \textit{TedEval (Text detection Evaluation)}, we evaluate text detections via an instance-level matching policy and a character-level scoring policy. Granularity is addressed by non-exclusively gathering all possible matches of one-to-one, one-to-many, and many-to-one. Afterwards, instance-level matches are scored while penalty is given to missing and overlapping characters. To this end, we use pseudo character centers made from word bounding boxes and their word lengths.

The main contributions of TedEval can be summarized as follows. 1) TedEval is simple, intuitive, and thus easy to use in a wide-range of tasks. 2) TedEval rewards behaviors that are favorable to recognition. 3) TedEval is relatively less affected by the quality of GT annotation. In this regard, we believe that TedEval can play a key role in developing state-of-the-art scene text detectors.

\section{Methodology}
Our evaluation metric performs the matching and the scoring of detectors through a separate policy. The matching process follows previous metrics, pairing bounding boxes of detections and GTs in an instance-level. On the other hand, the scoring process calculates recall and precision at the character-level without character annotations.

\begin{table}[b]
\vspace{-3mm}
    \fontsize{7.8}{7.8}\selectfont
    \renewcommand*{\arraystretch}{1.5}
    \centering
    \begin{tabular}{c|c||c|c|c|c|c||c}
    \hline
    \multicolumn{2}{c||}{} & \multicolumn{2}{c|}{ $D_1$ } & ... & \multicolumn{2}{c||}{ $D_j$ } & Recall \\
    \hline \hline
    \multirow{4}{*}{ $G_1$ } & $c_{1}^{1}$ & \multirow{4}{*}{ $M_{11}$ } & $m_{11}^{1}$ & \multirow{4}{*}{ } & \multirow{4}{*}{ $M_{1j}$ } & $m_{1j}^{1}$ & \multirow{4}{*}{$R_1$ } \\
    & $c_{1}^{2}$ & & $m_{11}^{2}$ & & & $m_{1j}^{2}$ & \\
    & ... & & ... & & & ... & \\
    & $c_{1}^{l_1}$ & & $m_{11}^{l_1}$ & & & $m_{1j}^{l_1}$ & \\
    \hline
    ... & ... & ... & ... & & ... & ... & ... \\
    \hline
    \multirow{4}{*}{ $G_i$ } & $c_{i}^{1}$ & \multirow{4}{*}{ $M_{i1}$ } & $m_{i1}^{1}$ & \multirow{4}{*}{ } & \multirow{4}{*}{ $M_{ij}$ } & $m_{ij}^{1}$ & \multirow{4}{*}{$R_i$ } \\
    & $c_{i}^{2}$ & & $m_{i1}^{2}$ & & & $m_{ij}^{2}$ & \\
    & ... & & ... & & & ... & \\
    & $c_{i}^{l_i}$ & & $m_{i1}^{l_i}$ & & & $m_{ij}^{l_i}$ & \\
    \hline \hline
    \multicolumn{2}{c||}{ Precision } & \multicolumn{2}{c|}{$P_1$ } &... & \multicolumn{2}{c||}{$P_j$ } & \\
    \hline
    \end{tabular}
    \caption{Visualization of the matching table with notation.}
    \label{tab:match}
\end{table}

\subsection{Matching process} \label{matching process}
Adopted from DetEval, our evaluation protocol measures area recall and area precision to pair GTs and detections in the manner of not only one-to-one but also one-to-many and many-to-one. To prevent side effects of addressing granularity, TedEval implements three major changes including (1) non-exclusive matching, (2) change in the area recall threshold, and (3) multiline prevention.

The first change directly relates to granularity. Accounting for granularity opens the possibility that a single instance (from either GT or detection) may satisfy multiple matches. DetEval assumes that the first match is the best match, discarding subsequent matches. This causes mismatch among one-to-one and many matches. A simple solution is to neither prioritize nor discard redundant matches. We accept all viable matches, setting $M_{ij}$ of the binary matrix from Table~\ref{tab:match} as 1 whenever a match between an instance of GT $G_i$ and detection $D_j$ occurs. When redundant matches occur, $M_{ij}$ is overwritten not accumulated.

Note that the threshold for area recall changed from 0.8 to 0.4. This reflects the interaction between the matching process and the scoring process introduced in the next section. Since the latter penalizes the incompleteness of detection to the given GT, namely recall, a lenient threshold in the matching process is fair.

Although our instance matching policy works in a fairly generous way, there is a case when the match itself should be prevented. Multiline is a special case of many match involving multiple lines of texts and must be rejected by identifying it in the matching stage. It involves not only many-to-one match where one detection contains multiple lines but also one-to-many match where multiple detections of different lines are matched to one GT.

\begin{figure}[t]
  \centering
  \includegraphics*[width=.5\linewidth, height=2.5cm, clip=true]{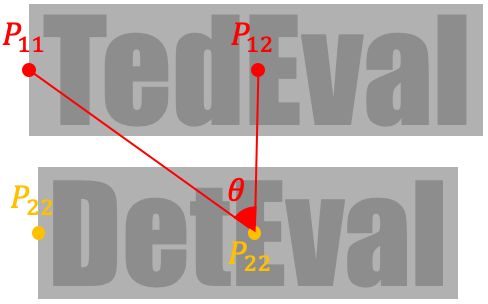}
  \caption{Visualization of multiline computation. The angle $\theta$ satisfies Eq.~\ref{eq:ml-angle} and thus this match is rejected.}
  \label{fig:ml}
  \vspace{-3mm}
\end{figure}

As shown in Fig.~\ref{fig:ml}, we use angles to identify multiline. Firstly, define $B$ as a set of bounding boxes in the many side of a many match. Then we compute two pivotal points for all bounding boxes in $B$ by:
\begin{equation} \label{eq:ml-char}
\begin{gathered}
	{p_{i1}} = mean(v_{i1}, v_{i4}) \\
    {p_{i2}} = mean(v_{i1}, v_{i2}, v_{i3}, v_{i4})
\end{gathered}
\end{equation}
where $v_{ij}$ means the $j$th vertex $(x,y)$ of $B_i$. Note that GT must assume the form of four vertices in a clockwise order starting from the top left point in regard to the orientation of the word.

The angle $\theta$ between $B_1$ and $B_2$ is then computed by turning from $p_{11}$ to $p_{12}$ around  $p_{22}$. While computing every possible angle among $B$, we reject the match when any one of it is:
\begin{equation} \label{eq:ml-angle}
    \left | min(\theta ,180-\theta ) \right |\geq 45^{\circ}.
\end{equation}
The threshold is obtained experimentally.

In addition, instead of selecting pivotal points as points on the edges, $p_{i1}$ is the point on the left edge and $p_{i2}$ is the center point of $B_i$. This is to make our algorithm robust against the width difference and distance between bounding boxes, which can confuse the magnitude of the angle. 

\subsection{Scoring process} \label{scoring process}
\begin{figure}[t]
    \centering
    \includegraphics[width=.7\linewidth]{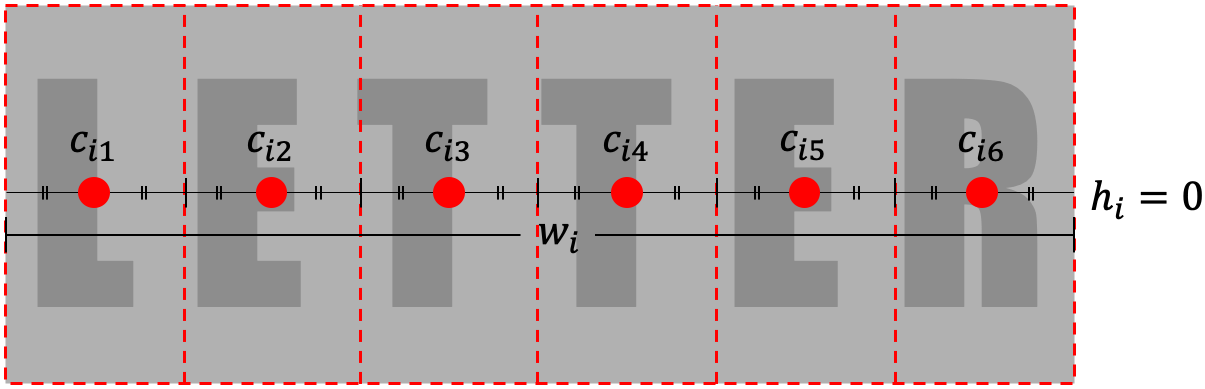}
    \caption{An example of computing PCC of $G_i$. Red dot: PCC. Red dash: pseudo character box. Grey: $G_i$.}
    \label{fig:missing-char}
    \vspace{-3mm}
\end{figure}

Based on instance-level matches from Section~\ref{matching process}, we calculate the recall score of $G_i$ and the precision score of $D_j$ in a character-level. To overcome the lack of character-level annotation in most public datasets, we compute Pseudo Character Centers (PCC) from word-level bounding boxes and their word lengths. As shown in Fig.~\ref{fig:missing-char}, a set of PCC of $G_i$ is computed by:
\begin{equation}
	\label{eq:pcc}
    c_i = \{ (x_i+\frac{w_i}{l_i}(k-\frac{1}{2}),y_i+\frac{h_i}{l_i}(k-\frac{1}{2}))\}_{k=1}^{l_i}
\end{equation}
where $x_i$ and $y_i$ are the x, y coordinates of $p_{i1}$ from Eq.~\ref{eq:ml-char}, $w_i$ and $h_i$ are $\Delta x$ and $\Delta y$ between $p_{i1}$ and the other edge, and $l_i$ is the word length. From the matching table in Table~\ref{tab:match}, $m_{ij}$ is a binary matrix whose element $m_{ij}^{k}$ is set to 1 when $D_j$ contains $c_{i}^{k}$.

For recall calculation, we perform a row-wise summation of character matching scores $m_{i}^{k}$:
\begin{equation} \label{eq:row-sum}
s_{i}^{k} = \sum_{j}{m_{ij}^{k}}.
\end{equation}
Since it is critical that each of the characters is detected only once, the condition of a correct character match is $s_{i}^{k}=1$. Contrastingly, mismatch cases include $s_{i}^{k}=0$ indicating missing characters and $s_{i}^{k}>1$ for overlap characters. The recall $R_{G_i}$ is the number of correct character matches over the text length $l_i$:
\begin{equation}
R_{G_{i}}=\frac{\left |\{ s_{i}^{k}=1 \}_{k=1}^{l_i} \right |}{l_i}.
\end{equation}
On the other hand, the precision $P_{D_j}$ is the number of correct character matches over the sum of text lengths of GTs matched with $D_j$:
\begin{equation} \label{eq:prec}
P_{D_{j}}=\frac{\sum_{i \in M_j}\sum_{k}{m_{ij}^{k}}}{\sum_{i \in M_j}{l_i}}\\
\end{equation}
where $M_j$ is $\{ x|M_{xj} = 1 \}$. Finally, $Recall$ and $Precision$ can be obtained by
\begin{equation}
	\begin{gathered}
        Recall = \frac{ \sum_{i=1}^{\left| G\right|}{R_{G_{i}}} }{\left| G\right|}, \\
        Precision = \frac{ \sum_{j=1}^{\left| D\right|}{P_{D_{j}}} }{\left| D\right|}.
    \end{gathered}
\end{equation}
Examples of our scoring process are in the Appendix.

Since scoring occurs column- and row-wise by instance, our scoring policy does not score the same instance multiple times even if it is involved in multiple matches. It can also differentiate a complete match from a partial match by penalizing missing or overlapping characters. This differs from instance-based scoring, which gives a binary score that does not reflect the completeness of detections.

In addition, TedEval automatically penalizes one-to-many cases, which may abuse the scoring policy by splitting a single word with multiple detections. For example, if a group of detections detect characters of GT once and completely, the precision score of each detection is the number of characters each detected over the length of the GT transcription. Then, the overall precision is 1 over the number of splits. Yet, penalty is not given to many-to-one cases. Examples of scoring many matches are in the Appendix.

\begin{table*}[t!]
    \fontsize{9}{9}\selectfont
    \renewcommand*{\arraystretch}{1.1}
    \centering
    \begin{tabular}{c||c|c|c|c|c|c|c||c|c|c|c|c|c|c}
        \hline
        \rule{0pt}{9pt} \multirow{3}{*}{ \textbf{Detector} } & \multicolumn{7}{c||}{ \textbf{ICDAR2013} } & \multicolumn{7}{c}{ \textbf{ICDAR2015} } \\
        \cline{2-15}
        \rule{0pt}{9pt} & \multicolumn{3}{c|}{ \textbf{DetEval} } & \multicolumn{3}{c|}{ \textbf{TedEval} } & \multirow{2}{*}{ \textbf{$\Delta$} } &
        \multicolumn{3}{c|}{ \textbf{IoU} } & \multicolumn{3}{c|}{ \textbf{TedEval} } & \multirow{2}{*}{ \textbf{$\Delta$} } \\
        \cline{2-7} \cline{9-14}
        \rule{0pt}{9pt} & \textbf{R} & \textbf{P} & \textbf{H} & \textbf{R} & \textbf{P} & \textbf{H} & & \textbf{R} & \textbf{P} & \textbf{H} & \textbf{R} & \textbf{P} & \textbf{H} & \\
        \hline \hline
        SegLink~\cite{shi2017seglink} & 60.0 & 73.9 & 66.2 & 65.6 & 74.9 & 70.0 & \textcolor{red}{3.8} & 72.9 & 80.2 & 76.4 & 77.1 & 83.9 & 80.6 & \textcolor{red}{4.2} \\
        EAST~\cite{zhou2017east} & 70.7 & 81.6 & 75.8 & 77.7 & 87.1 & 82.5 & \textcolor{red}{6.7} & 77.2 & 84.6 & 80.8 & 82.5 & 90.0 & 86.3 & \textcolor{red}{5.5} \\
        CTPN~\cite{tian2016ctpn} & 83.0 & 93.0 & 87.7 & 82.1 & 92.7 & 87.6 & \textcolor{blue}{-0.1} & 51.6 & 74.2 & 60.9 & 85.0 & 81.1 & 67.8 & \textcolor{red}{6.9} \\
        PixelLink~\cite{deng2018pixellink} & 87.5 & 88.7 & 88.1 & 84.0 & 87.2 & 86.1 & \textcolor{blue}{-2.0} & 83.8 & 86.7 & 85.2 & 85.7 & 86.1 & 86.0 & \textcolor{red}{0.8} \\
        TextBoxes++~\cite{liao2018textboxes++} & 85.6 & 91.9 & 88.6 & 87.4 & 92.3 & 90.0 & \textcolor{red}{1.4} & 80.8 & 89.1 & 84.8 & 82.4 & 90.8 & 86.5 & \textcolor{red}{1.7} \\
        WordSup~\cite{hu2017wordsup} & 87.1 & 92.8 & 89.9 & 87.5 & 92.2 & 90.2 & \textcolor{red}{0.3} & 77.3 & 80.5 & 78.9 & 83.2 & 87.1 & 85.2 & \textcolor{red}{6.3} \\
        RRPN~\cite{ma2017rrpn} & 87.3 & 95.2 & 91.1 & 89.0 & 94.2 & 91.6 & \textcolor{red}{0.5} & 77.1 & 83.5 & 80.2 & 79.5 & 85.9 & 82.6 & \textcolor{red}{2.4} \\
        MaskTextSpotter~\cite{lyu2018mask} & 88.6 & 95.0 & 91.7 & 90.2 & 95.4 & 92.9 & \textcolor{red}{1.2} & 79.5 & 89.0 & 84.0 & 82.5 & 91.8 & 86.9 & \textcolor{red}{2.9} \\
        FOTS~\cite{liu2018fots} & 90.4 & 95.4 & 92.8 & 91.5 & 93.0 & 92.6 & \textcolor{blue}{-0.2} & 87.9 & 91.9 & 89.8 & 89.0 & 93.4 & 91.2 & \textcolor{red}{1.4} \\
        PMTD~\cite{liu2019pmtd} & 92.2 & 95.1 & 93.6 & 94.0 & 95.2 & 94.7 & \textcolor{red}{1.1} & 87.4 & 91.3 & 89.3 & 89.2 & 92.8 & 91.0 & \textcolor{red}{1.7} \\
        CRAFT~\cite{baek2019craft} & 93.1 & 97.4 & 95.2 & 93.6 & 96.5 & 95.1 & \textcolor{blue}{-0.1} & 84.3 & 89.8 & 86.9 & 88.5 & 93.1 & 90.9 & \textcolor{red}{4.0} \\
        \hline
    \end{tabular}
    \caption{Comparison of evaluation metrics for different detectors. R, P, and H refer to recall, precision, and H-mean. Detectors are sorted from the highest score on DetEval metric. Texts are highlighted in \textcolor{red}{red} and \textcolor{blue}{blue} for rise and fall.}
    \label{tab:result}
    \vspace{-3mm}
\end{table*}

\begin{figure*}[t!]
    \begin{subfigure}{.5\linewidth}
    \centering
    \includegraphics[width=\linewidth, height=4.5cm]{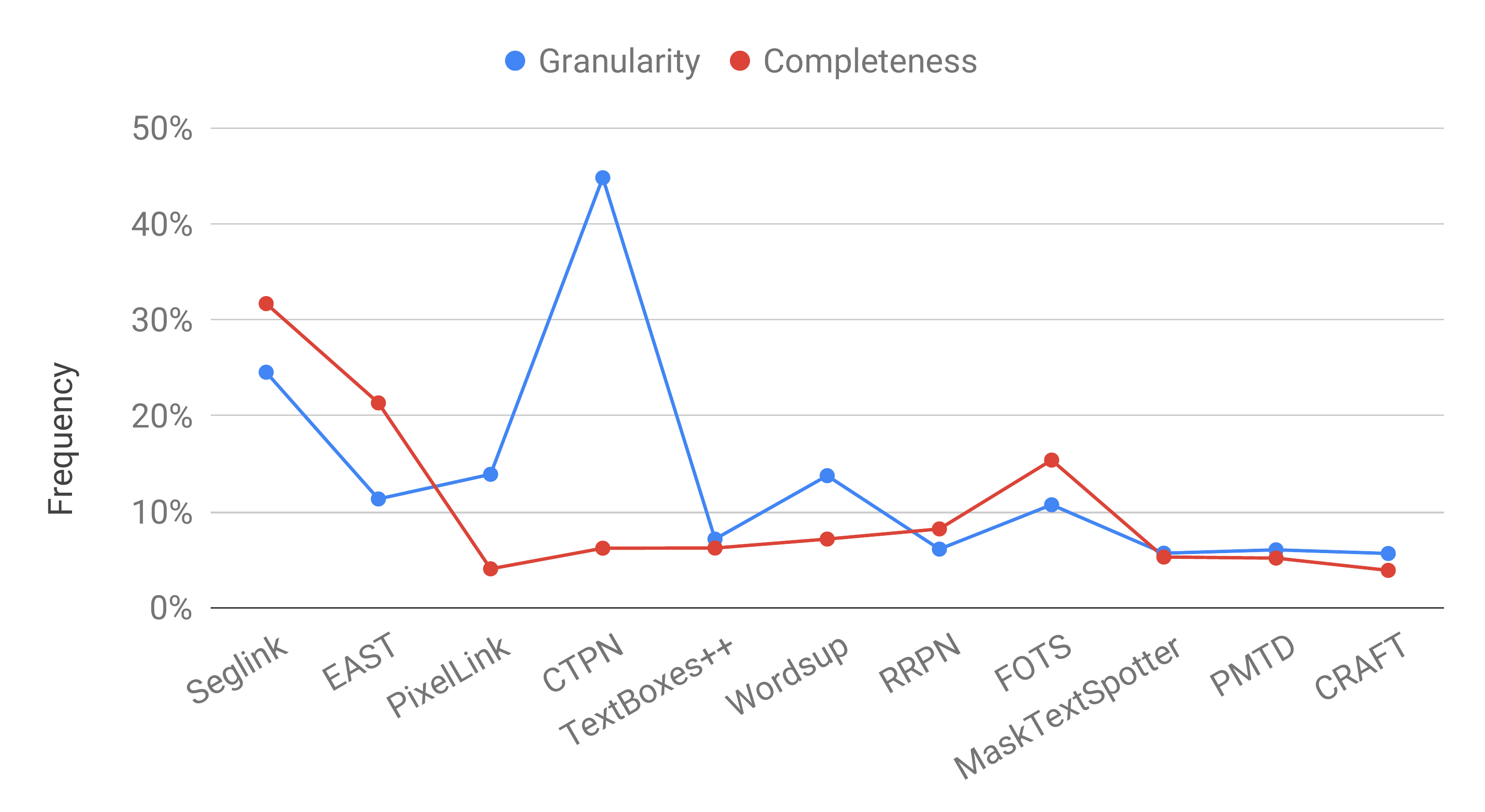}
    \caption{IC13}
    \label{fig:count-ic13}
    \end{subfigure}%
    \begin{subfigure}{.5\linewidth}
    \centering
    \includegraphics[width=\linewidth, height=4.5cm]{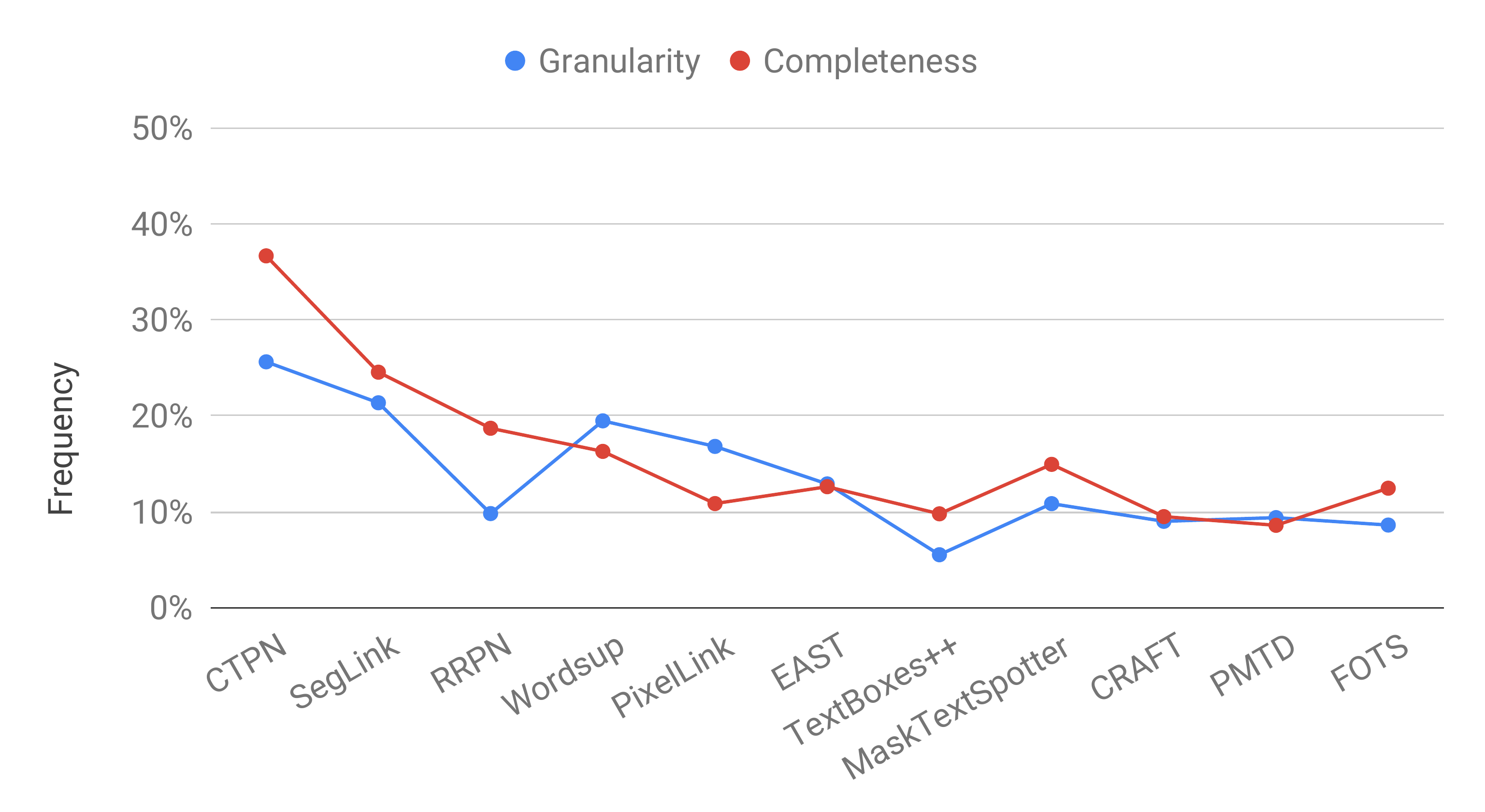}
    \caption{IC15}
    \label{fig:count-ic15}
    \end{subfigure}
    \vspace{-3mm}
    \caption{Frequency of factors that TedEval tackles counted by predictions. Numbers are represented as proportions to the number of successful detections. Detectors are sorted right to left from the highest score on TedEval.}
    \label{fig:count}
\end{figure*}

\begin{figure*}[t!]
    \begin{subfigure}{.25\linewidth}
    \centering
    \includegraphics[width=\linewidth, height=2cm]{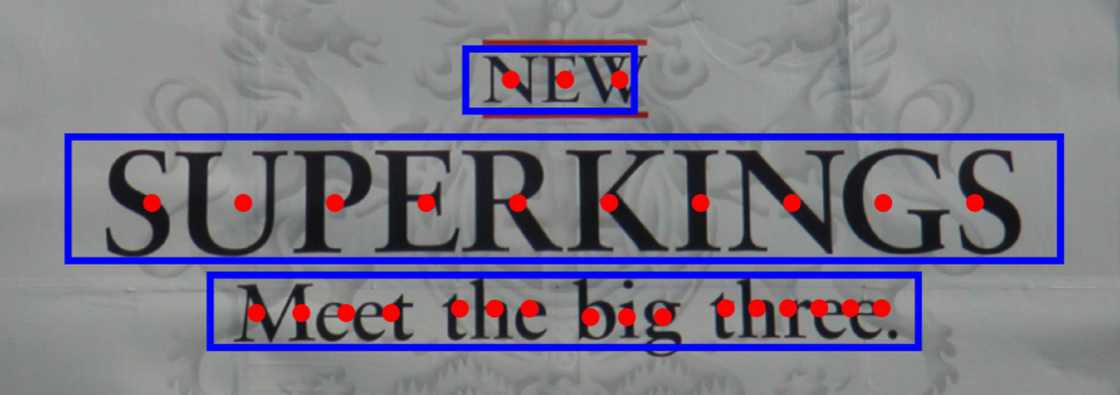}
    \caption{CTPN ($R:1.00$, $P:1.00$)}
    \end{subfigure}%
    \begin{subfigure}{.25\linewidth}
    \centering
    \includegraphics[width=\linewidth, height=2cm]{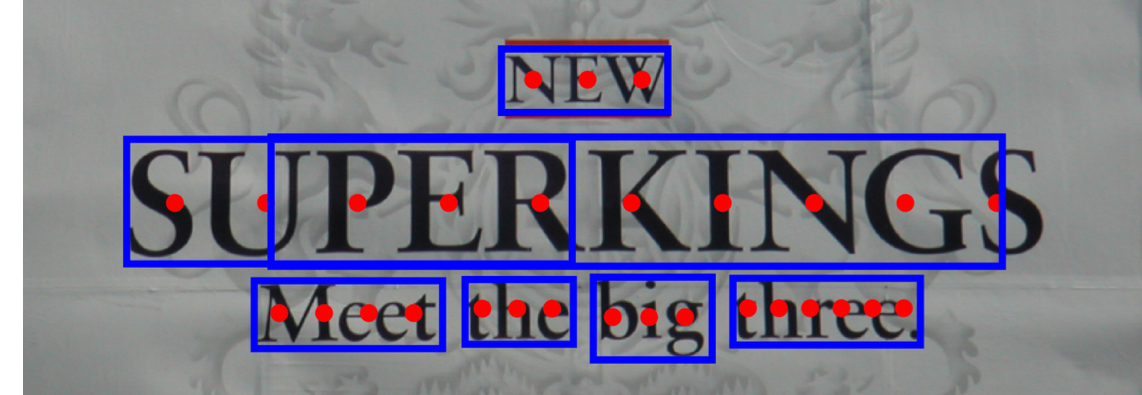}
    \caption{FOTS ($R:0.70$, $P:0.65$)}
    \end{subfigure}%
    \begin{subfigure}{.25\linewidth}
    \centering
    \includegraphics[width=\linewidth, height=2cm]{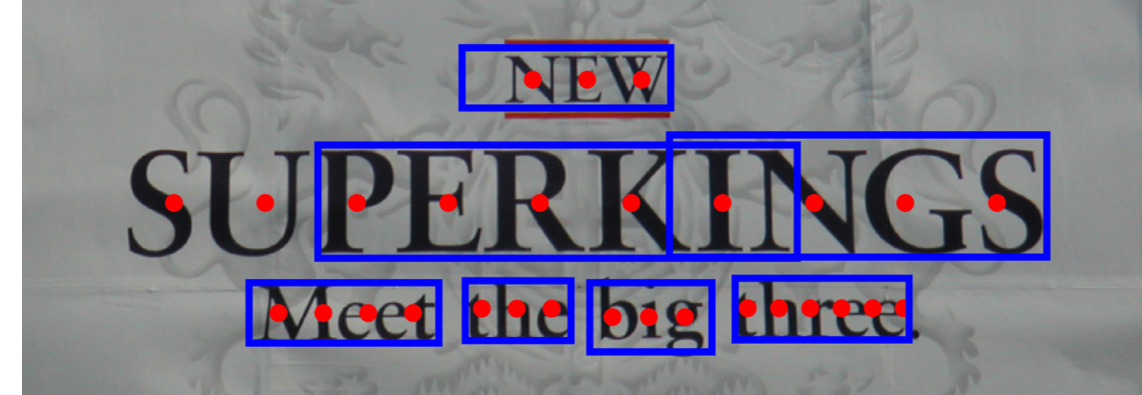}
    \caption{EAST ($R:0.70$, $P:0.45$)}
    \end{subfigure}%
    \begin{subfigure}{.25\linewidth}
    \centering
    \includegraphics[width=\linewidth, height=2cm]{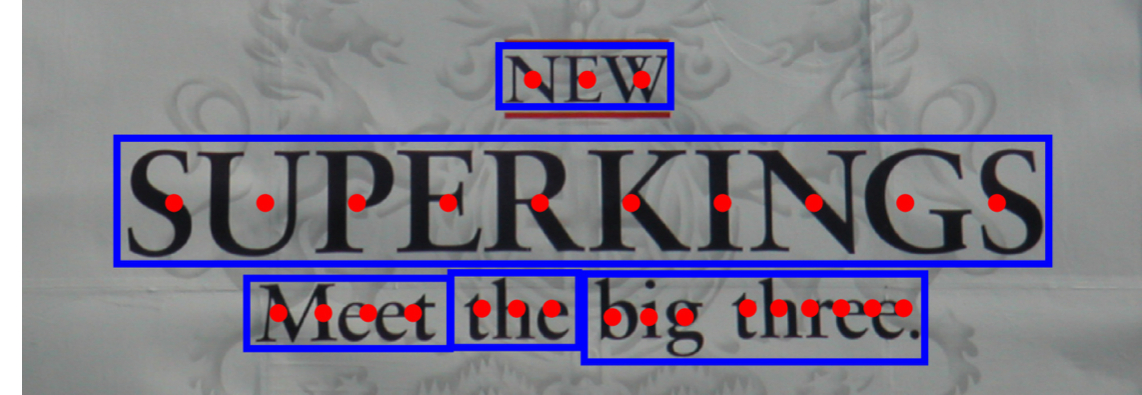}
    \caption{PixelLink ($R:1.00$, $P:1.00$)}
    \end{subfigure}
\vspace{-3mm}
\caption{Examples of incomplete detections. Numbers in caption indicate recall and precision scores of "SUPERKINGS." Red dot: PCC. Blue: detection.}
\label{fig:example}
\vspace{-3mm}
\end{figure*}
\section{Experiments}
We compared TedEval with DetEval and IoU on two public datasets: ICDAR 2013 Focused Scene Text (IC13) and ICDAR 2015 Incidental Scene Text (IC15). We requested from authors the result file of scene text detectors that frequently appear and are referenced in the literature. Results are shown in Table~\ref{tab:result}.

Fig.~\ref{fig:count} shows the frequency of factors that TedEval tackles as proportions to the number of successful detections. Detectors are sorted right to left from the highest score on TedEval. Granularity counts in average 14\% and 14\% and completeness counts in average 10\% and 16\% in IC13 and IC15, respectively. These proportions considerably influence the change of H-mean scores $\Delta$ in Table \ref{tab:result} and are the main causes of qualitative discords in previous metrics.

Delving into some of the peaks in Fig.~\ref{fig:count}, notice that CTPN has the highest granularity frequency in both datasets. As shown in Fig.~\ref{fig:example}, CTPN has a tendency to detect a single box for an entire line, namely many-to-one. Since such behaviors are not penalized by TedEval, CTPN gains a 6.8 increase in the H-mean score in IC15 from IoU, which does not account for granularity.

Another peak is from FOTS. As shown in Fig.~\ref{fig:example}, FOTS often detects a word by splitting it into several parts while causing overlap between such detections. This causes peaks for both granularity and completeness and lowers the recall score. Note that EAST, which proposed the detector architecture of FOTS, shows similar behaviors.

On the contrary, PixelLink and CRAFT have noticeably low completeness counts. They are both segmentation-based detections, which perform well in finding text regions. However, since they connect text regions using link information, they often detect multiline in a single box. In fact, the multiline proportion of PixelLink and CRAFT are one of the highest, 7\% and 3\%, respectively.

More examples can be seen in the Appendix.
\section{Conclusion}
We have proposed a novel evaluation metric for scene text detectors called TedEval, which evaluates text detections by an instance-level matching policy and a character-level scoring policy. It accounts for granularity by adopting DetEval but implements a few changes to prevent subsequent side effects of many matches. A scoring policy uses pseudo character centers to reflect the penalty given to missing and overlapping characters to the final recall and precision score.

Experiments on two public datasets demonstrated that issues TedEval tackles frequently occur in results from state-of-the-art detectors and that they caused qualitative disagreements in previous metrics. By reflecting such factors, TedEval can provide a fair and reliable evaluation on the state-of-the-art methods in the upper percentile of H-mean scores.

Our future work involves evaluating polygon annotations, where TedEval's logics would be more effective, and making our logic insensitive to the vertex order. This will make TedEval easier to apply in various tasks and lessen the burden of annotators.

\vspace{2mm}
\noindent\textbf{Acknowledgements.} We would like to thank Yuliang Liu, the author of TIoU, and authors of scene text detectors who kindly provided result files used in our experiments.

{\small
\bibliographystyle{ieee}
\bibliography{mybib}
}

\onecolumn
\clearpage
\appendix
\section{Matching matrix}

\begin{figure*}[h]
  \centering
  \includegraphics*[height=20cm, width=.8\linewidth, clip=true]{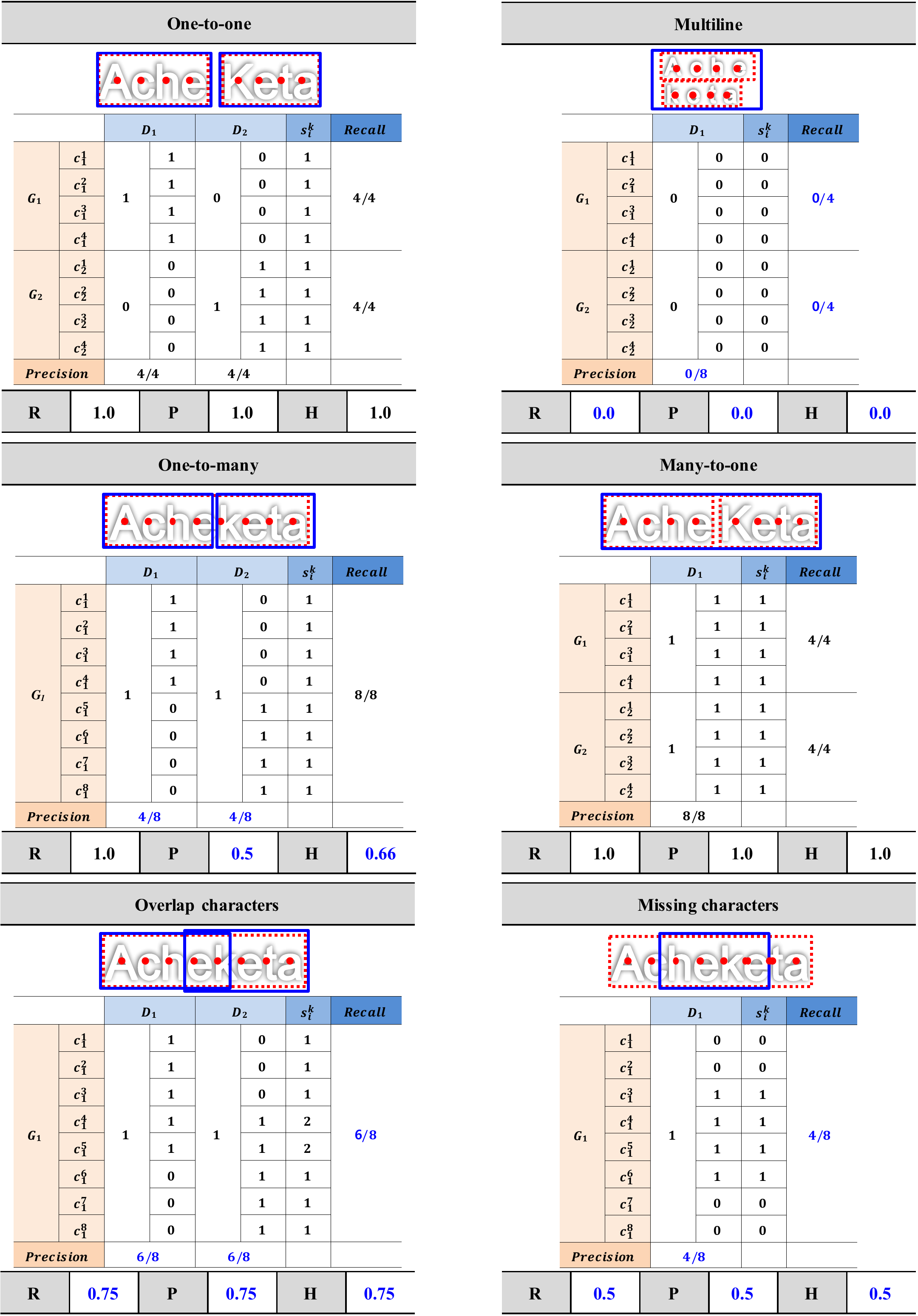}
  \caption{Examples of scoring in various cases.}
  \label{fig:ex-score} 
\end{figure*}

\clearpage
\section{Detection results}

\begin{figure*}[h]
    \begin{subfigure}{.25\linewidth}
    \centering
    \includegraphics[width=\linewidth, height=3.5cm]{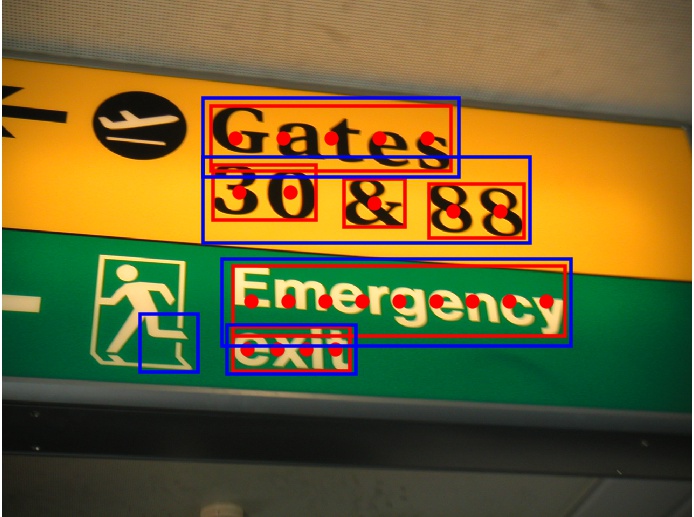}
    \caption{PixelLink ($R:1.00$, $P:0.80$)}
    \end{subfigure}%
    \begin{subfigure}{.25\linewidth}
    \centering
    \includegraphics[width=\linewidth, height=3.5cm]{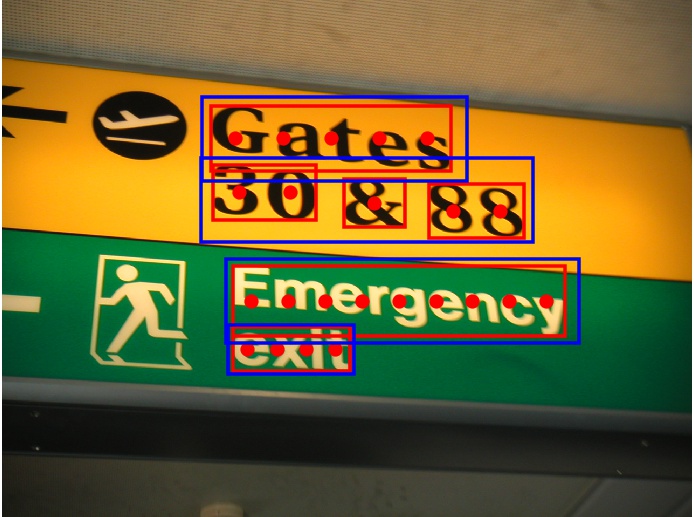}
    \caption{WordSup ($R:1.00$, $P:1.00$)}
    \end{subfigure}%
    \begin{subfigure}{.25\linewidth}
    \centering
    \includegraphics[width=\linewidth, height=3.5cm]{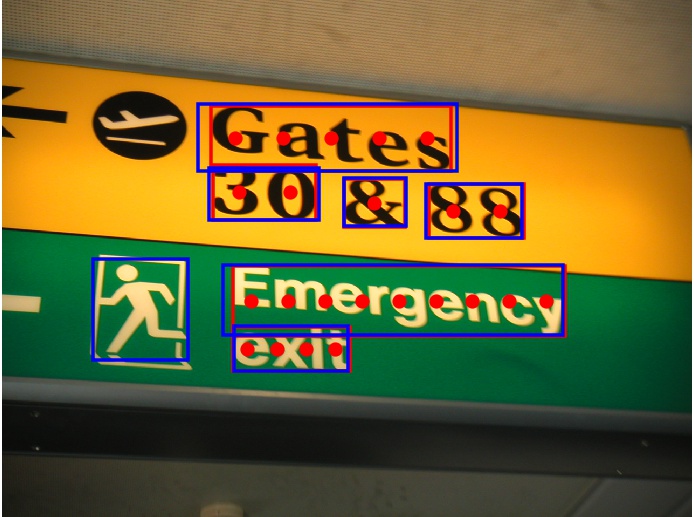}
    \caption{TB++ ($R:1.00$, $P:0.86$)}
    \end{subfigure}%
    \begin{subfigure}{.25\linewidth}
    \centering
    \includegraphics[width=\linewidth, height=3.5cm]{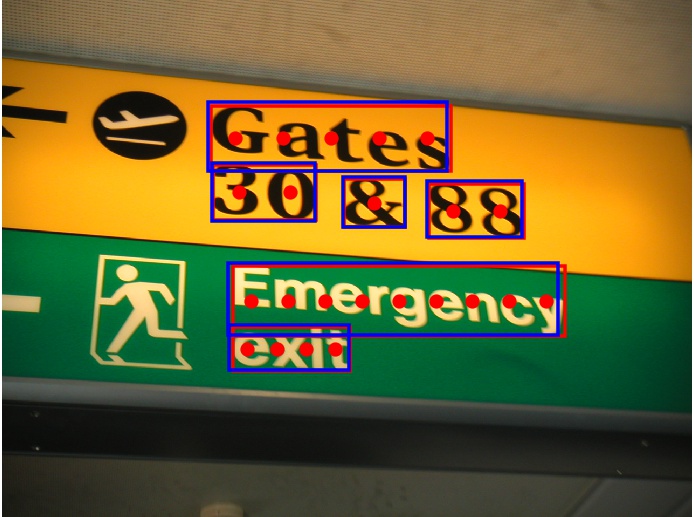}
    \caption{PMTD ($R:1.00$, $P:1.00$)}
    \end{subfigure}
\vspace{-3mm}
\caption{Granularity}
\end{figure*}

\begin{figure*}[h]
    \begin{subfigure}{.25\linewidth}
    \centering
    \includegraphics[width=\linewidth, height=3.5cm]{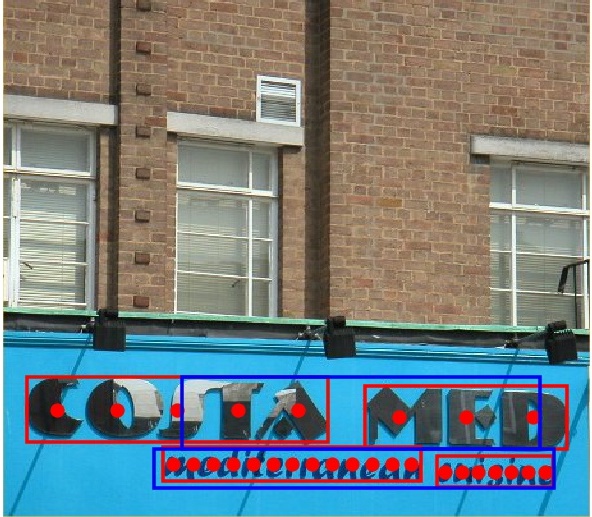}
    \caption{CTPN ($R:0.75$, $P:1.00$)}
    \end{subfigure}%
    \begin{subfigure}{.25\linewidth}
    \centering
    \includegraphics[width=\linewidth, height=3.5cm]{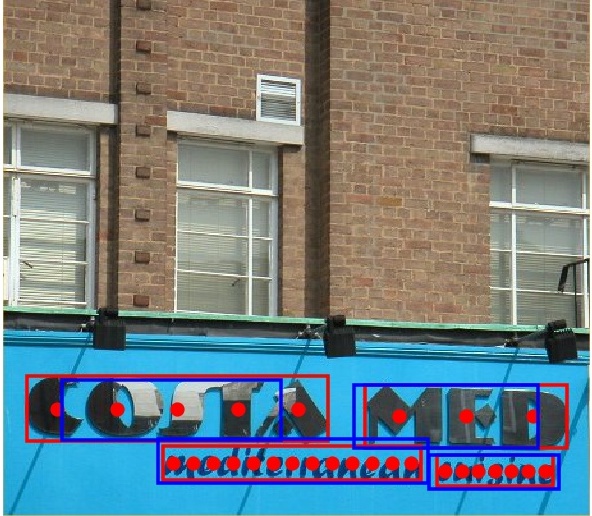}
    \caption{PixelLink ($R:0.90$, $P:0.90$)}
    \end{subfigure}%
    \begin{subfigure}{.25\linewidth}
    \centering
    \includegraphics[width=\linewidth, height=3.5cm]{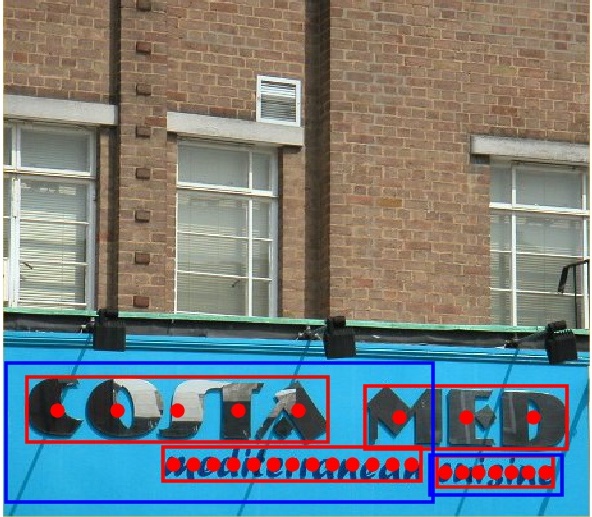}
    \caption{WordSup ($R:0.25$, $P:0.50$)}
    \end{subfigure}%
    \begin{subfigure}{.25\linewidth}
    \centering
    \includegraphics[width=\linewidth, height=3.5cm]{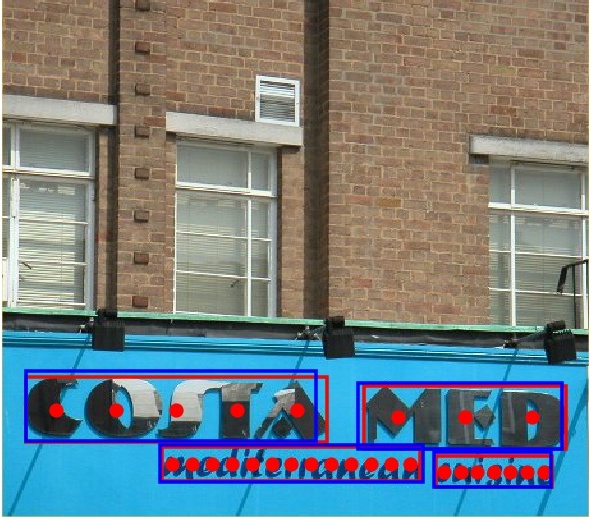}
    \caption{MaskTS ($R:1.00$, $P:1.00$)}
    \end{subfigure}
\vspace{-3mm}
\caption{Completeness}
\end{figure*}

\begin{figure*}[h]
    \begin{subfigure}{.25\linewidth}
    \centering
    \includegraphics[width=\linewidth, height=3.5cm]{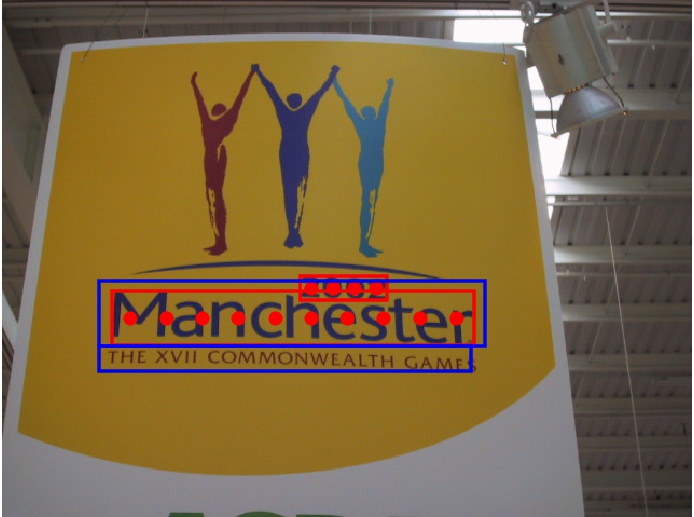}
    \caption{CTPN ($R:0.83$, $P:1.00$)}
    \end{subfigure}%
    \begin{subfigure}{.25\linewidth}
    \centering
    \includegraphics[width=\linewidth, height=3.5cm]{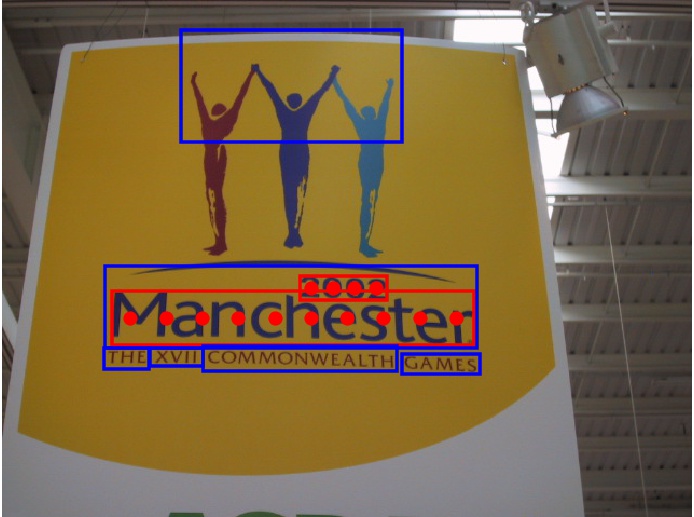}
    \caption{PixelLink ($R:0.83$, $P:0.83$)}
    \end{subfigure}%
    \begin{subfigure}{.25\linewidth}
    \centering
    \includegraphics[width=\linewidth, height=3.5cm]{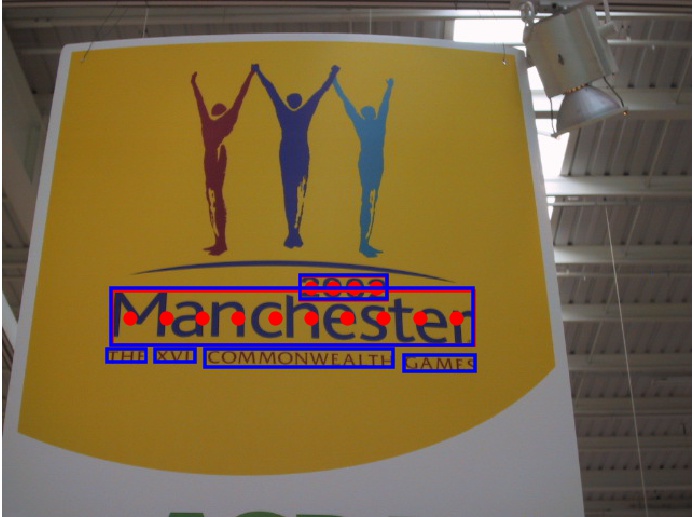}
    \caption{MaskTS ($R:1.00$, $P:1.00$)}
    \end{subfigure}%
    \begin{subfigure}{.25\linewidth}
    \centering
    \includegraphics[width=\linewidth, height=3.5cm]{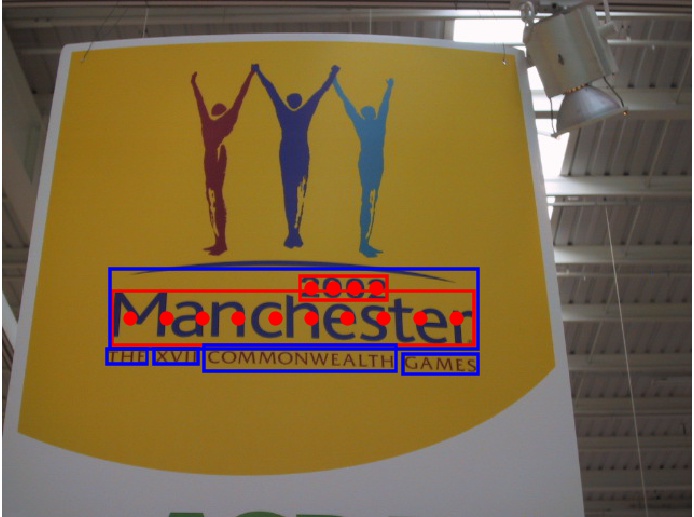}
    \caption{CRAFT ($R:0.83$, $P:1.00$)}
    \end{subfigure}
\vspace{-3mm}
\caption{Multiline}
\end{figure*}

\begin{figure*}[h]
    \begin{subfigure}{.25\linewidth}
    \centering
    \includegraphics[width=\linewidth, height=3.5cm]{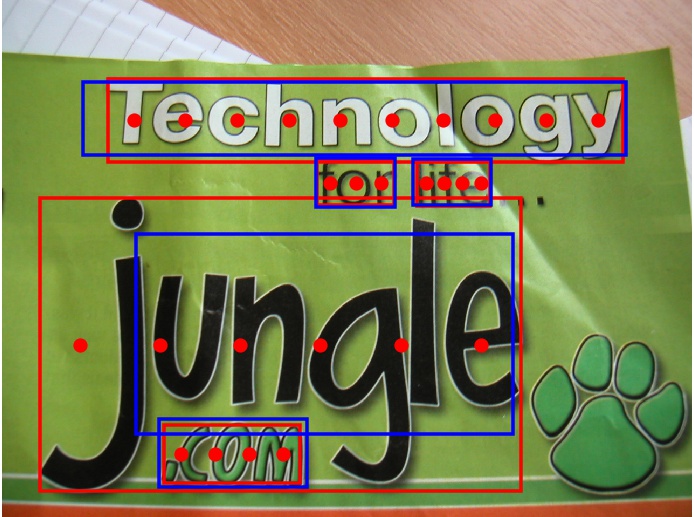}
    \caption{TB++ ($R:0.97$, $P:0.97$)}
    \end{subfigure}%
    \begin{subfigure}{.25\linewidth}
    \centering
    \includegraphics[width=\linewidth, height=3.5cm]{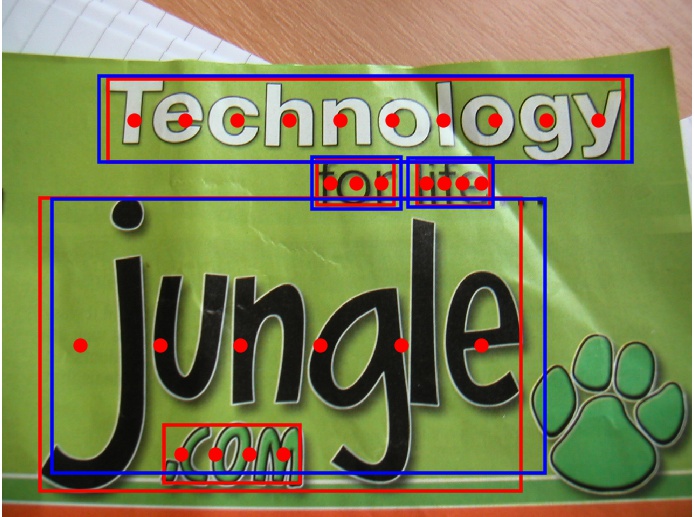}
    \caption{WordSup ($R:0.80$, $P:1.00$)}
    \end{subfigure}%
    \begin{subfigure}{.25\linewidth}
    \centering
    \includegraphics[width=\linewidth, height=3.5cm]{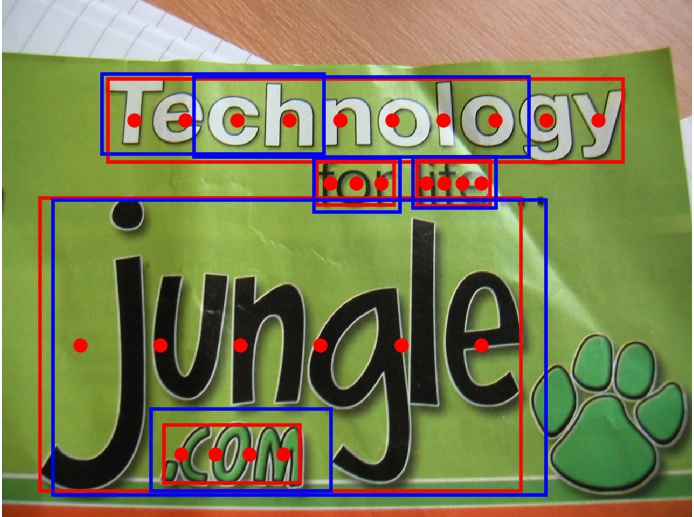}
    \caption{FOTS ($R:0.92$, $P:0.83$)}
    \end{subfigure}%
    \begin{subfigure}{.25\linewidth}
    \centering
    \includegraphics[width=\linewidth, height=3.5cm]{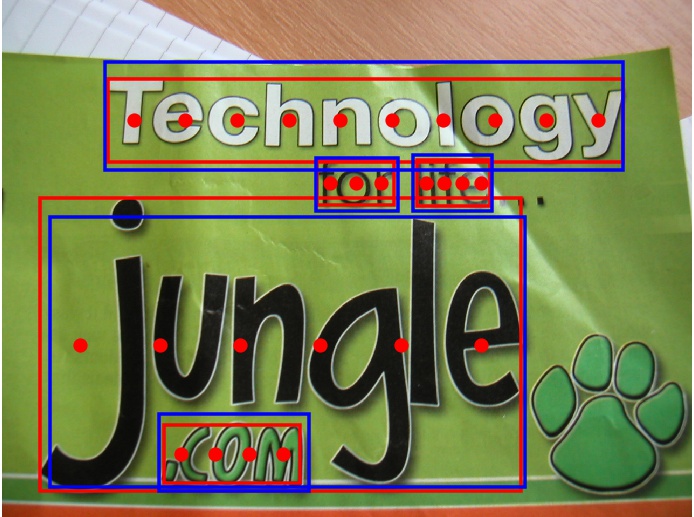}
    \caption{CRAFT ($R:1.00$, $P:1.00$)}
    \end{subfigure}
\vspace{-3mm}
\caption{Text-in-text}
\end{figure*}

\end{document}